%% file: main.tex
\pdfoutput=1
\documentclass[runningheads,final]{llncs}

\makeatletter
\def\mdseries@tt{m}
\def\mdseries@rm{m}
\makeatother

\usepackage[T1]{fontenc}
\usepackage[utf8]{inputenc}

\usepackage[usenames,dvipsnames]{xcolor}
\usepackage{algorithmic}
\usepackage{amsmath}
\usepackage{array}
\usepackage[pdftex]{graphicx}
\usepackage{placeins}
\usepackage[obeyFinal,prependcaption]{todonotes}
\usepackage{changes}
\usepackage{url}
\usepackage{xspace}
\usepackage{tikz}
\usetikzlibrary{bending,positioning,calc,patterns,backgrounds,fit,decorations.pathreplacing}
\usetikzlibrary{shapes.misc,shapes}
\usetikzlibrary{decorations.pathreplacing}
\usetikzlibrary{arrows.meta,decorations.pathreplacing,fadings,shapes,arrows}
\usepackage{ifdraft}
\usepackage{ifthen}

\usepackage{dblfloatfix}
\usepackage{paralist}
\usepackage{booktabs}

\usepackage[style=lncs,sorting=none,minbibnames=3,natbib=true,url=false]{biblatex}
\usepackage[normalem]{ulem}

\usepackage{caption}
\DeclareCaptionFont{white}{\color{white}}
\DeclareCaptionFormat{listing}{\colorbox{gray}{\parbox{\linewidth}{#1#2#3}}}
\usepackage{subcaption}

\usepackage{listings,lstautogobble}
\usepackage[newfloat,finalizecache=false,frozencache=true]{minted}
\setminted{fontsize=\footnotesize}
\setminted{breaklines=true}
\setminted{autogobble=true}
\setminted{frame=bottomline}
\setminted{fontfamily=courier}

\usepackage{hyperref}

\newcommand\myshade{85}
\colorlet{mylinkcolor}{violet}
\colorlet{mycitecolor}{YellowOrange}
\colorlet{myurlcolor}{Aquamarine}

\hypersetup{
  linkcolor  = mylinkcolor!\myshade!black,
  citecolor  = mycitecolor!\myshade!black,
  urlcolor   = myurlcolor!\myshade!black,
  colorlinks = true,
  breaklinks = true, 
}
\usepackage{breakurl} 

\usepackage{cleveref}

\usepackage{xfrac}
\usepackage[detect-all,binary-units]{siunitx}
\newcommand{\todoforJWW}[2][]{\todo[fancyline,color=green,#1]{#2 (for JWW)}}

\newcommand{\BSS}[1]{BSS\nobreakdash-#1}
\newcommand{\BrainScaleS}[1]{BrainScaleS\nobreakdash-#1}

\begin{document}

\title{Inference with Artificial Neural Networks on Analog Neuromorphic Hardware}
\titlerunning{Inference with ANNs on Analog Neuromorphic Hardware}

\author{%
		Johannes Weis
		\and
		Philipp Spilger
		\and
		Sebastian Billaudelle
		\and
		Yannik Stradmann
		\and
		Arne Emmel
		\and
		Eric Müller
		\and
		Oliver Breitwieser
		\and
        Andreas Grübl
        \and
        Joscha Ilmberger
        \and
        Vitali Karasenko
        \and
        Mitja Kleider
		\and
		Christian Mauch
        \and
        Korbinian Schreiber
        \and\\
		Johannes Schemmel%
}
\authorrunning{J.\ Weis et al.}

\institute{Kirchhoff-Institute for Physics\\
Ruprecht-Karls-Universität Heidelberg, Germany\\
\email{johannes.weis@kip.uni-heidelberg.de}
}

\maketitle


\begin{abstract}
\input{tex/0_abstract}
\end{abstract}


\input{tex/1_intro}
\input{tex/2_methods}
\input{tex/3_results}
\input{tex/4_discussion}
\input{tex/5_contributions}

\section*{Acknowledgments}

The authors wish to thank all present and former members of the Electronic Vision(s) research group contributing to the BrainScaleS-2 hardware platform, software development as well as operation methodologies.
We especially express our gratefulness to the late Karlheinz Meier who initiated and led the project for most if its time.

This work has received funding from the EU (%
[H2020/2014-2020]%
)
under grant agreements
720270,  
785907 and  
945539 (HBP) 
as well as from the BMBF (16ES1127 (HD-BIO-AI)).

\printbibliography[notkeyword=own_software]
\printbibliography[title={Own Software},keyword=own_software]

\end{document}

%% file: tex/0_abstract.tex
The neuromorphic \BrainScaleS{2} ASIC comprises mixed-sig\-nal neurons and synapse circuits as well as two versatile digital microprocessors.
Primarily designed to emulate spiking neural networks, the system can also operate in a vector-matrix multiplication and accumulation mode for artificial neural networks.
Analog multiplication is carried out in the synapse circuits, while the results are accumulated on the neurons' membrane capacitors.
Designed as an analog, in-memory computing device, it promises high energy efficiency.
Fixed-pattern noise and trial-to-trial variations, however, require the implemented networks to cope with a certain level of perturbations.
Further limitations are imposed by the digital resolution of the input values~(\mbox{5 bit}), matrix weights~(\mbox{6 bit}) and resulting neuron activations~(\mbox{8 bit}).
In this paper, we discuss \BrainScaleS{2} as an analog inference accelerator and present calibration as well as optimization strategies, highlighting the advantages of training with hardware in the loop.
Among other benchmarks, we classify the MNIST handwritten digits dataset using a two-dimensional convolution and two dense layers.
We reach \SI{98.0}{\percent} test accuracy, closely matching the performance of the same network evaluated in software.

\keywords{%
Analog Accelerator \and
Neural Network Processor \and
Neuromorphic Hardware \and
Convolutional Neural Networks \and
Machine Learning \and
In-memory Computing \and
MNIST%
}

%% file: tex/1_intro.tex
\section{Introduction}

Artificial neural networks (ANN) find application in a wide variety of fields and problems.
With networks growing in depth and complexity, the increase of computational cost becomes more and more significant \citep{brown2020language}.
In fact, execution time and power consumption often represent the crucial limiting factors in further scaling and in the application of ANNs \citep{schwartz2019green}.

A large fraction of the computational cost for neural network-based inference is spent on vector-matrix multiplications \citep{oh2004gpu}.
With their massive parallelization of floating point calculations, GPUs already cut runtime significantly compared to CPUs.
Computational complexity can often be cut by representing and processing data with reduced precision  \citep{micikevicius2017mixed}.
Specialized digital inference accelerators have been presented \citep{jouppi2017datacenter, yin20171} that offer further efficiency improvements over implementations on general-purpose hardware.
Potentially even more efficient ASICs could be based on mixed-signal circuit designs by exploiting physical processes for computational purposes \citep{boser_jssc91}.
Drawbacks of such systems can include vulnerability to fixed-pattern and trial-to-trial variations, resulting in distorted network configurations and reduced reproducibility.
Similar to  digital solutions with reduced precision, networks have to cope with limited weight resolution, which can be as low as one bit  \citep{yamaguchi2019energy}.

In this work, we demonstrate \BrainScaleS{2} \citep{schemmel2020accelerated} as an analog inference accelerator.
We describe the hardware configuration and operating principle of analog vector-matrix multiplication on the ASIC and benchmark the system's performance by training and classifying the MNIST dataset of handwritten digits \citep{lecunmnist}.
We further discuss calibration and the benefits of training with hardware in the loop as strategies to counter chip-specific fixed-pattern variations \citep{schmitt2017hwitl, cramer2020training}.

%% file: tex/2_methods.tex
\section{Methods}
\label{subsec:bss2accelerator}

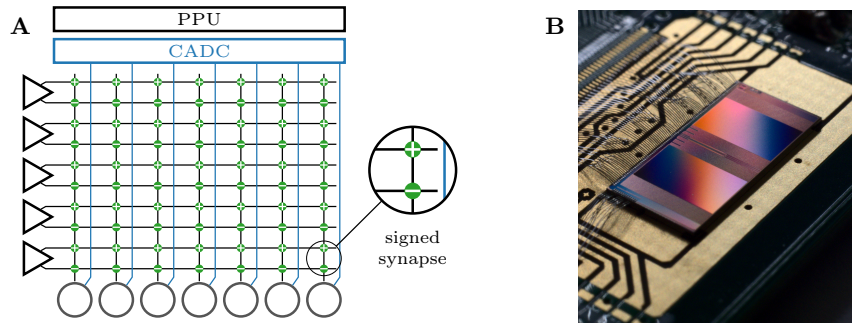
\begin{figure}[t]
    \begin{subfigure}{0.6\textwidth}
        \begin{tikzpicture}
            \draw (0, 0) node[anchor=north west, inner sep=0] {
                \scalebox{1.2}{
                    \hspace{-15em}
                    \input{tex/chip_layout.tikz}}
            };
            \node[inner sep=0pt] at (0.95, -0.3) {\textbf{A}};
        \end{tikzpicture}
    \end{subfigure}
    \hfill
    \begin{subfigure}{0.35\textwidth}
        \begin{tikzpicture}
            \draw (0, 0) node[anchor=north west, inner sep=0] {
                \includegraphics[trim=0 100 0 150, clip, width=0.85\textwidth]{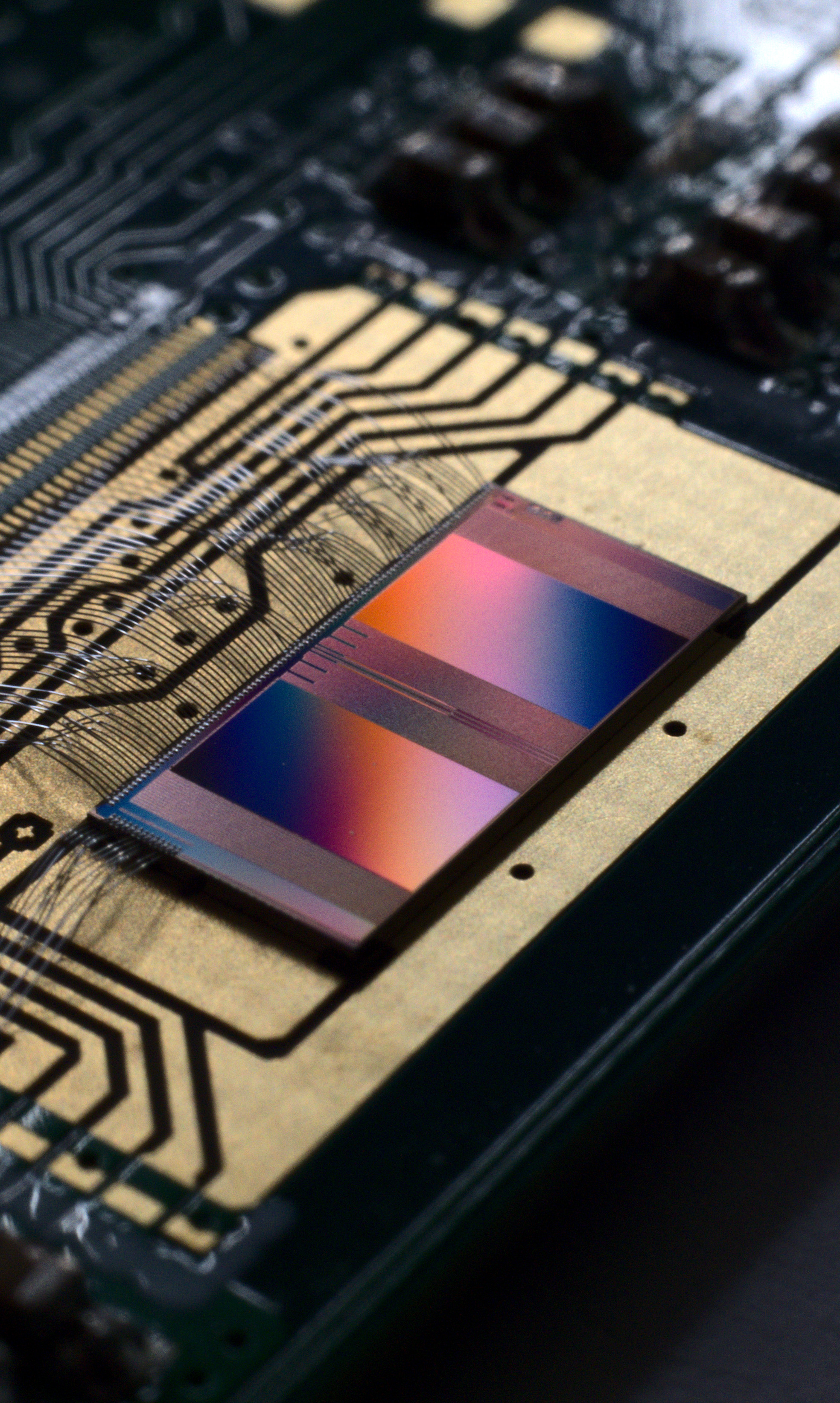}
            };
            \node[inner sep=0pt] at (-0.3, -0.2) {\textbf{B}};
        \end{tikzpicture}
    \end{subfigure}
    \caption{
        Overview of the \BrainScaleS{2} system.
        \textbf{A}: Block diagram of the analog core, showing synapse drivers (triangles), neurons (large circles), and synapses (small circles in matrix).
        Signed weights are achieved by using two synapse rows for positive and negative weights.
        Figure adopted from \citep{cramer2020training}.
        \textbf{B}: Chip photograph.
    }
        \label{fig:hx_layout}
\end{figure}

\begin{figure}[t]
    \begin{subfigure}{0.48\textwidth}
        \begin{tikzpicture}
            \draw (0, 0) node[anchor=north west, inner sep=0] {
                \scalebox{0.38}{
                    \includegraphics{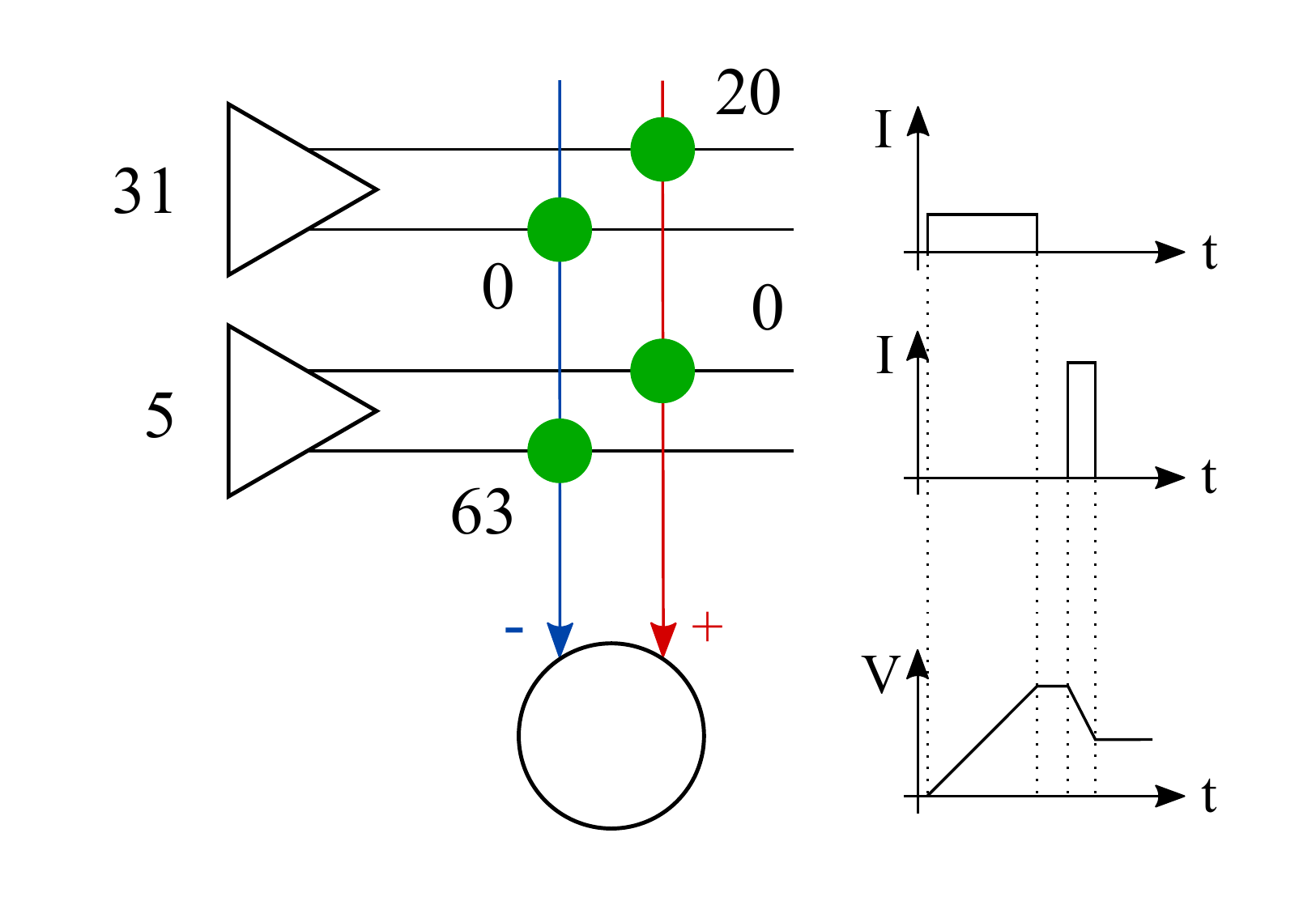}}
            };
            \node[inner sep=0pt] at (0.7, -0.4) {\textbf{A}};
        \end{tikzpicture}
    \end{subfigure}
    \hfill
    \begin{subfigure}{0.48\textwidth}
        \begin{tikzpicture}
            \draw (0, 0) node[anchor=north west, inner sep=0] {
                \input{plots/neuron_integration.pgf}
            };
            \node[inner sep=0pt] at (0.6, -0.5) {\textbf{B}};
        \end{tikzpicture}
    \end{subfigure}
    \caption{
        \textbf{A}: Illustration of a multiply-accumulate operation.
        The vector value controls the length of the current pulses, the matrix weight their amplitude.
        The currents are integrated on the neuron's membrane.
        \textbf{B}: The recorded membrane trace clearly shows the integration phase where the synaptic inputs are integrated, after which the result is digitized (dotted line).
        Afterwards, the voltage decays exponentially towards the resting potential.
    }
    \label{fig:hx_mac_example}
\end{figure}
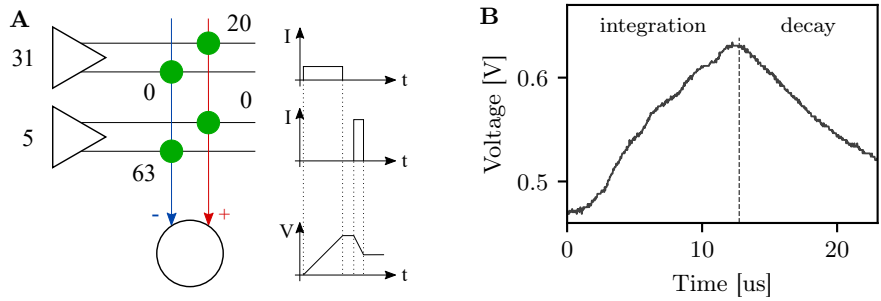

\BrainScaleS{2} is a mixed-signal ASIC fabricated in a \SI{65}{\nano\meter} CMOS process by TSMC that has originally been designed as an accelerator for biologically plausible spiking neural networks.
It features analog circuits emulating neurons and synapses as well as digital periphery for communication, parameter storage, and realtime control.
Recent additions to the system allow for in-memory computation of multiply-accumulate operations within the chip's analog core, thereby making the system applicable for inference with artificial neural networks \citep{schemmel2020accelerated}.
Matrix multiplication can also be combined with spiking operation, thus seamless integration of a partially spiking network is possible.
The chip contains \num{512} analog neurons arranged in two blocks, each neuron receives input from \num{256} synapses.
Thus, \BrainScaleS{2} can be used to multiply a vector with \num{256} entries to a matrix comprising \num{512} columns.
An architectural overview of the circuitry for processing vector-matrix multiplications on \BrainScaleS{2} is depicted in \cref{fig:hx_layout}.
Digitally encoded input vectors are injected from the left, converted to the analog domain and multiplied within the central synapse array.
Each neuron accumulates values from its corresponding synapse column.
The resulting vector of neuron activations is read out in parallel via a columnar analog-to-digital converter (CADC)\@.

\paragraph{Multiplication in analog synapse circuits.}
Within the synapse array, the multiplication of an input value with the synaptic weight is modelled as the electrical charge $Q = I \cdot \Delta t$ emitted during a current pulse of variable length and amplitude.
The current $I$ is determined by a 6 bit weight stored locally in each synapse.
The time window $\Delta t$ during which that current is emitted is modulated by circuitry in the synapse drivers (triangles on the left in \cref{fig:hx_layout}).
The value is set by the payload of input events, which is otherwise used to select a subset of synapses from a row.
More specifically, we use 5 bit of this label to encode the pulse length $\Delta t$.

Each row of synapses can be connected to the afferent neurons with either positive or negative sign.
To achieve signed weights, two rows of synapses can be combined to represent a single logical row.
This configuration, however, reduces the number of available vector entries from \num{256} to \num{128}.
When using signed weights, the remaining input label bit is still available to differentiate two sets of synapses, therefore two different multiplications can be executed side by side.

\paragraph{Neurons integrate synaptic currents.}
Each neuron uses its membrane to accumulate the individual multiplication results from its respective synaptic column.
They integrate the positive and negative charge contributions, as sketched in \cref{fig:hx_mac_example}A.
Motivated by spiking operation, the input signals are low-pass filtered with finite time constant.
To speed up integration and for reduction of synaptic input saturation, the minimum time constant of approximately \SI{1}{\micro\second} was configured.

The neurons' dynamics are based on a leaky integrator model commonly used for spiking networks \citep{brette_05}:
A resistor continuously pulls the membrane voltage, which is physically represented across a capacitor, towards a resting potential.
The resulting dynamics constitute a crucial part for the emulation of spiking networks.
In contrast, they can lead to distortions in the accumulation of vector matrix multiplication results which do not contain implicit timing information.
In order to stabilize the accumulated voltages and reduce the effect of noise, we configured a rather large but finite resistance.
The leak resistance leads to an exponential decay of the integrated charge, which can be seen in \cref{fig:hx_mac_example}B.

\paragraph{Digitization of results.}
The membrane potentials are digitized in parallel for all \num{256} neurons connected to a synapse matrix, using the CADCs, and stored via the on-chip microprocessors.
The resulting 8 bit values represent the neuron activations and are the result of the multiply-accumulate (MAC) operation.

By aligning the choice of the resting potential with the dynamic range of the ADC, two operating modes can be selected:
In case the lower end of the ADC range coincides with the resting potential, negative activations are cut off.
In this configuration, the neurons behave as hardware rectified linear units (ReLU).

In case the inputs a neuron receives exceed the size of the synapse matrix, the network can be partitioned into smaller matrices which are evaluated in a time multiplexed fashion \citep{spilger2020hxtorch}.
Since the activation function needs to be applied after combining the individual results, negative activations must be representable.
For this purpose, the resting potential can be chosen centered in the ADCs' dynamic ranges.

\subsection{Structure of a multiply-accumulate operation}\label{sec:single-mac}

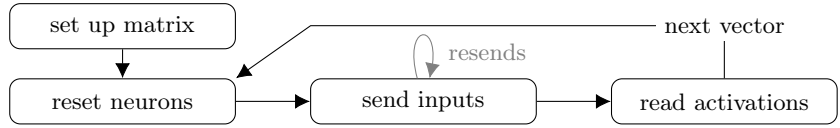
\begin{figure}[t]
    \begin{center}
        \input{tex/flowchart_mac.tikz}
    \end{center}
    \caption{Pattern for executing multiply-accumulate operations between a matrix and a batch of vectors.}
    \label{fig:flowchart_mac}
\end{figure}

To compile a MAC operation from the elements outlined above, the sequence shown in \cref{fig:flowchart_mac} can be applied.

\begin{enumerate}
    \item To begin with, the weight matrix is written to the synapses.
    Writing all \num{256}\texttimes\num{512} values takes about \SI{5}{\milli\second}.
    Batched execution can minimize the amount of expensive reconfiguration.
    \item  A reset of the membrane potentials removes any previous state accumulated by the neurons.
    An immediate read of the voltages establishes a baseline activation to suppress low frequency noise.
    Resetting all neurons takes approximately \SI{1}{\micro\second}.
    \item The vector inputs are sent sequentially to the chip.
    Between events, wait times of \SIrange{8}{200}{\nano\second} are inserted.
    These mitigate saturation effects in the neurons' synaptic inputs, which can occur in case multiple inputs of large amplitude are sent in a short period of time.
    To improve the signal-to-noise ratio, the activations on the membranes can be increased by incorporating resends of the input vectors within a single integration phase.
    Wait times as well as the number of resends must be optimized considering the neurons' decay times, which limit the maximum integration time.
    Alternatively, the membrane capacitance can be reduced, yielding higher activations, but also shorter decay time constants.
    Skipping inputs of zero reduces the overall runtime, especially in conjunction with ReLU activation functions.
    \item The activations are digitized after the accumulation of charges.
    Considering the finite time constant of the synaptic inputs, a waiting period of \SI{2}{\micro\second} is inserted for the membrane potential to settle.
    The ADC conversion takes \SI{1.5}{\micro\second}.
\end{enumerate}

\subsection{Calibration}

Transistor-level mismatch in the manufacturing process of an ASIC leads to inhomogeneous electrical properties of the fabricated circuits.
Due to the analog nature of \BrainScaleS{2}, the resulting fixed-pattern variations cause each neuron and synapse to behave differently when presented with similar input.
Without calibration, networks sensitive to such perturbations can not perform up to their full potential on the analog substrate, as weights and activations would be distorted.
\BrainScaleS{2} therefore provides a substantial amount of digitally controlled parameters that allow equalization of all computational units through calibration.

The operating point of the neuron circuits is determined by a set of internal parameters and references.
Some parameters are of technical nature and need to be calibrated to ensure correct operation.
Others directly influence the circuit dynamics, e.g.\ the neurons' time constants and resting potentials.
Most importantly, the strength of the synaptic currents has to be equalized across neurons:
It determines the increment of the membrane potential as a response to the synaptic stimulation.
The response of all \num{512} neurons to constant stimuli is shown in the histograms of \cref{fig:characterization}A,\ separately for the calibrated and uncalibrated states.
For both, positive and negative contributions, the synaptic strength has been calibrated to a precision of \SI{7}{\%}.

Calibration is also applied to the pulse generation circuits in the synapse drivers.
The length of the current pulses encoding the 5 bit input values is subject to row-wise random offsets.
Calibration registers allow to strongly reduce these variations to \SI{0.3}{\nano\second}.

We developed a collection of optimization routines for \BSS{2}, which automate such calibration steps and allow to set the hardware up for matrix multiplication usage.
The code is based on the Python API of \citet{mueller2020bss2ll} and may also be used for configuring the system for spiking neural networks.

\subsection{Training with hardware in the loop}

Remaining imperfections after calibration can still lead to a loss of performance when directly transferring trained network models to an analog computing substrate.
In the context of spiking neural networks, it has been previously shown that integrating the analog hardware into the training loop can restore the original performance \citep{schmitt2017hwitl, cramer2020training}.
For some parameters, such training in the loop can in fact replace explicit calibration \citep{wunderlich2019demonstrating}.
For the analog matrix multiplication described in this manuscript, the gradients are calculated based on measured activations, only assuming linearity of the synaptic weights.
A detailed description of the implementation and integration into PyTorch \citep{paszke2019pytorch} is given by \citet{spilger2020hxtorch}.

%% file: tex/chip_layout.tikz
\usetikzlibrary{spy}%
\definecolor{blue}{HTML}{1f77b4}%
\definecolor{red}{HTML}{d62728}%
\definecolor{green}{HTML}{2ca02c}%
\definecolor{orange}{HTML}{ff710e}%
\definecolor{yellow}{HTML}{fee23e}%
\definecolor{cadc}{HTML}{1f77b4}%
\definecolor{input}{HTML}{ff7f0e}%
\definecolor{hidden}{HTML}{2ca02c}%
\definecolor{output}{HTML}{555555}%
\tikzset{block/.style={font={\rmfamily\footnotesize},align=center}}%
\tikzset{box/.style={draw=black!90}}%
\tikzset{block label/.style={fill=white,font={\rmfamily\footnotesize},inner sep=0.05cm}}%
\tikzset{%
	neuron/.style = {%
		draw=black,%
		circle,%
		inner sep=0pt,%
		minimum width=0.4cm%
	},%
	driver/.style = {%
		minimum height=0.45cm,%
		draw=black,%
		regular polygon,%
		regular polygon sides=3,%
		shape border rotate=-90,%
		inner sep=0pt%
	},%
}%
\begin{tikzpicture}[
            scale=0.8,
            >=stealth,
            transform shape,
	    line width=1.0\pgflinewidth,
	    anchor=center,
	    spy using outlines=circle,
        ]
        \pgfdeclarelayer{background layer}
        \pgfsetlayers{background layer,main}
        \draw[use as bounding box,inner sep=0pt,draw=none] (0.0,0.0) rectangle ++(7.0,4.5);

	\begin{scope}[scale=1.15]
		\foreach \x in {0,1,...,6} {

			\node[neuron,output,thick] (nrn \x) at (0.8 + \x*0.5,0.35) {};
			\draw (nrn \x.north) ++ (0.0,0.01) -- ++(0.0,2.5);

		}

		\foreach \y in {0,1,...,4} {
			\node[driver,thick] (drv \y) at ($(nrn 0) + (-0.5,0.5 + \y*0.5)$) {};
			\draw (drv \y.+45) -- ($(drv \y.center) + (0.15, 0.125)$) -- ++(3.5,0.0);
			\draw (drv \y.-45) -- ($(drv \y.center) + (0.15,-0.125)$) -- ++(3.5,0.0);

			\foreach \x in {0,1,...,6} {

				\fill[hidden] (drv \y -| nrn \x) ++ (0.0, 0.125) circle (0.05cm);
				\draw[white] (drv \y -| nrn \x) ++ (0.0, 0.125) ++ (-0.045,0.0) -- ++(0.09,0.0);
				\draw[white] (drv \y -| nrn \x) ++ (0.0, 0.125) ++ (0.0,-0.045) -- ++(0.0,0.09);
				\fill[hidden] (drv \y -| nrn \x) ++ (0.0,-0.125) circle (0.05cm);
				\draw[white] (drv \y -| nrn \x) ++ (0.0,-0.125) ++ (-0.045,0.0) -- ++(0.09,0.0);
			}
		}

		\node[rectangle,thick,draw,cadc,inner sep=2pt,minimum width=3.5cm] (cadc) at ($(nrn 2) + (0.5,3.0)$) {\fontsize{6}{6}\selectfont CADC};
		\foreach \x in {0,1,...,6} {
			\draw[blue] (nrn \x.55) -- ++(55:0.12) coordinate (tmp) -- (tmp |- cadc.south);
		}
		
		\node[rectangle,thick,draw,inner sep=2pt,minimum width=3.5cm,above=0.07cm of cadc] (ppu) {\fontsize{6}{6}\selectfont PPU};
		
	\end{scope}

	\coordinate (sherlock) at (5.6,2.2);
	\spy[draw,height=1.2cm,width=1.2cm,magnification=2.5,connect spies] on (nrn 6 |- drv 0) in node at (sherlock);
	\node[align=center] at ($(sherlock) - (0.0,1.1)$) {\fontsize{7}{7}\selectfont signed\\[-0.3em] \fontsize{7}{7}\selectfont synapse};
\end{tikzpicture}

%% file: plots/neuron_integration.pgf
\begingroup%
\makeatletter%
\begin{pgfpicture}%
\pgfpathrectangle{\pgfpointorigin}{\pgfqpoint{2.400000in}{1.800000in}}%
\pgfusepath{use as bounding box, clip}%
\begin{pgfscope}%
\pgfsetbuttcap%
\pgfsetmiterjoin%
\definecolor{currentfill}{rgb}{1.000000,1.000000,1.000000}%
\pgfsetfillcolor{currentfill}%
\pgfsetlinewidth{0.000000pt}%
\definecolor{currentstroke}{rgb}{1.000000,1.000000,1.000000}%
\pgfsetstrokecolor{currentstroke}%
\pgfsetdash{}{0pt}%
\pgfpathmoveto{\pgfqpoint{0.000000in}{0.000000in}}%
\pgfpathlineto{\pgfqpoint{2.400000in}{0.000000in}}%
\pgfpathlineto{\pgfqpoint{2.400000in}{1.800000in}}%
\pgfpathlineto{\pgfqpoint{0.000000in}{1.800000in}}%
\pgfpathclose%
\pgfusepath{fill}%
\end{pgfscope}%
\begin{pgfscope}%
\pgfsetbuttcap%
\pgfsetmiterjoin%
\definecolor{currentfill}{rgb}{1.000000,1.000000,1.000000}%
\pgfsetfillcolor{currentfill}%
\pgfsetlinewidth{0.000000pt}%
\definecolor{currentstroke}{rgb}{0.000000,0.000000,0.000000}%
\pgfsetstrokecolor{currentstroke}%
\pgfsetstrokeopacity{0.000000}%
\pgfsetdash{}{0pt}%
\pgfpathmoveto{\pgfqpoint{0.639851in}{0.523007in}}%
\pgfpathlineto{\pgfqpoint{2.265000in}{0.523007in}}%
\pgfpathlineto{\pgfqpoint{2.265000in}{1.665000in}}%
\pgfpathlineto{\pgfqpoint{0.639851in}{1.665000in}}%
\pgfpathclose%
\pgfusepath{fill}%
\end{pgfscope}%
\begin{pgfscope}%
\pgfsetbuttcap%
\pgfsetroundjoin%
\definecolor{currentfill}{rgb}{0.000000,0.000000,0.000000}%
\pgfsetfillcolor{currentfill}%
\pgfsetlinewidth{0.803000pt}%
\definecolor{currentstroke}{rgb}{0.000000,0.000000,0.000000}%
\pgfsetstrokecolor{currentstroke}%
\pgfsetdash{}{0pt}%
\pgfsys@defobject{currentmarker}{\pgfqpoint{0.000000in}{-0.048611in}}{\pgfqpoint{0.000000in}{0.000000in}}{%
\pgfpathmoveto{\pgfqpoint{0.000000in}{0.000000in}}%
\pgfpathlineto{\pgfqpoint{0.000000in}{-0.048611in}}%
\pgfusepath{stroke,fill}%
}%
\begin{pgfscope}%
\pgfsys@transformshift{0.639851in}{0.523007in}%
\pgfsys@useobject{currentmarker}{}%
\end{pgfscope}%
\end{pgfscope}%
\begin{pgfscope}%
\definecolor{textcolor}{rgb}{0.000000,0.000000,0.000000}%
\pgfsetstrokecolor{textcolor}%
\pgfsetfillcolor{textcolor}%
\pgftext[x=0.639851in,y=0.425785in,,top]{\color{textcolor}\rmfamily\fontsize{9.000000}{10.800000}\selectfont \(\displaystyle 0\)}%
\end{pgfscope}%
\begin{pgfscope}%
\pgfsetbuttcap%
\pgfsetroundjoin%
\definecolor{currentfill}{rgb}{0.000000,0.000000,0.000000}%
\pgfsetfillcolor{currentfill}%
\pgfsetlinewidth{0.803000pt}%
\definecolor{currentstroke}{rgb}{0.000000,0.000000,0.000000}%
\pgfsetstrokecolor{currentstroke}%
\pgfsetdash{}{0pt}%
\pgfsys@defobject{currentmarker}{\pgfqpoint{0.000000in}{-0.048611in}}{\pgfqpoint{0.000000in}{0.000000in}}{%
\pgfpathmoveto{\pgfqpoint{0.000000in}{0.000000in}}%
\pgfpathlineto{\pgfqpoint{0.000000in}{-0.048611in}}%
\pgfusepath{stroke,fill}%
}%
\begin{pgfscope}%
\pgfsys@transformshift{1.346438in}{0.523007in}%
\pgfsys@useobject{currentmarker}{}%
\end{pgfscope}%
\end{pgfscope}%
\begin{pgfscope}%
\definecolor{textcolor}{rgb}{0.000000,0.000000,0.000000}%
\pgfsetstrokecolor{textcolor}%
\pgfsetfillcolor{textcolor}%
\pgftext[x=1.346438in,y=0.425785in,,top]{\color{textcolor}\rmfamily\fontsize{9.000000}{10.800000}\selectfont \(\displaystyle 10\)}%
\end{pgfscope}%
\begin{pgfscope}%
\pgfsetbuttcap%
\pgfsetroundjoin%
\definecolor{currentfill}{rgb}{0.000000,0.000000,0.000000}%
\pgfsetfillcolor{currentfill}%
\pgfsetlinewidth{0.803000pt}%
\definecolor{currentstroke}{rgb}{0.000000,0.000000,0.000000}%
\pgfsetstrokecolor{currentstroke}%
\pgfsetdash{}{0pt}%
\pgfsys@defobject{currentmarker}{\pgfqpoint{0.000000in}{-0.048611in}}{\pgfqpoint{0.000000in}{0.000000in}}{%
\pgfpathmoveto{\pgfqpoint{0.000000in}{0.000000in}}%
\pgfpathlineto{\pgfqpoint{0.000000in}{-0.048611in}}%
\pgfusepath{stroke,fill}%
}%
\begin{pgfscope}%
\pgfsys@transformshift{2.053024in}{0.523007in}%
\pgfsys@useobject{currentmarker}{}%
\end{pgfscope}%
\end{pgfscope}%
\begin{pgfscope}%
\definecolor{textcolor}{rgb}{0.000000,0.000000,0.000000}%
\pgfsetstrokecolor{textcolor}%
\pgfsetfillcolor{textcolor}%
\pgftext[x=2.053024in,y=0.425785in,,top]{\color{textcolor}\rmfamily\fontsize{9.000000}{10.800000}\selectfont \(\displaystyle 20\)}%
\end{pgfscope}%
\begin{pgfscope}%
\definecolor{textcolor}{rgb}{0.000000,0.000000,0.000000}%
\pgfsetstrokecolor{textcolor}%
\pgfsetfillcolor{textcolor}%
\pgftext[x=1.452426in,y=0.259229in,,top]{\color{textcolor}\rmfamily\fontsize{9.000000}{10.800000}\selectfont Time [us]}%
\end{pgfscope}%
\begin{pgfscope}%
\pgfsetbuttcap%
\pgfsetroundjoin%
\definecolor{currentfill}{rgb}{0.000000,0.000000,0.000000}%
\pgfsetfillcolor{currentfill}%
\pgfsetlinewidth{0.803000pt}%
\definecolor{currentstroke}{rgb}{0.000000,0.000000,0.000000}%
\pgfsetstrokecolor{currentstroke}%
\pgfsetdash{}{0pt}%
\pgfsys@defobject{currentmarker}{\pgfqpoint{-0.048611in}{0.000000in}}{\pgfqpoint{0.000000in}{0.000000in}}{%
\pgfpathmoveto{\pgfqpoint{0.000000in}{0.000000in}}%
\pgfpathlineto{\pgfqpoint{-0.048611in}{0.000000in}}%
\pgfusepath{stroke,fill}%
}%
\begin{pgfscope}%
\pgfsys@transformshift{0.639851in}{0.740529in}%
\pgfsys@useobject{currentmarker}{}%
\end{pgfscope}%
\end{pgfscope}%
\begin{pgfscope}%
\definecolor{textcolor}{rgb}{0.000000,0.000000,0.000000}%
\pgfsetstrokecolor{textcolor}%
\pgfsetfillcolor{textcolor}%
\pgftext[x=0.378471in, y=0.697154in, left, base]{\color{textcolor}\rmfamily\fontsize{9.000000}{10.800000}\selectfont \(\displaystyle 0.5\)}%
\end{pgfscope}%
\begin{pgfscope}%
\pgfsetbuttcap%
\pgfsetroundjoin%
\definecolor{currentfill}{rgb}{0.000000,0.000000,0.000000}%
\pgfsetfillcolor{currentfill}%
\pgfsetlinewidth{0.803000pt}%
\definecolor{currentstroke}{rgb}{0.000000,0.000000,0.000000}%
\pgfsetstrokecolor{currentstroke}%
\pgfsetdash{}{0pt}%
\pgfsys@defobject{currentmarker}{\pgfqpoint{-0.048611in}{0.000000in}}{\pgfqpoint{0.000000in}{0.000000in}}{%
\pgfpathmoveto{\pgfqpoint{0.000000in}{0.000000in}}%
\pgfpathlineto{\pgfqpoint{-0.048611in}{0.000000in}}%
\pgfusepath{stroke,fill}%
}%
\begin{pgfscope}%
\pgfsys@transformshift{0.639851in}{1.284336in}%
\pgfsys@useobject{currentmarker}{}%
\end{pgfscope}%
\end{pgfscope}%
\begin{pgfscope}%
\definecolor{textcolor}{rgb}{0.000000,0.000000,0.000000}%
\pgfsetstrokecolor{textcolor}%
\pgfsetfillcolor{textcolor}%
\pgftext[x=0.378471in, y=1.240961in, left, base]{\color{textcolor}\rmfamily\fontsize{9.000000}{10.800000}\selectfont \(\displaystyle 0.6\)}%
\end{pgfscope}%
\begin{pgfscope}%
\definecolor{textcolor}{rgb}{0.000000,0.000000,0.000000}%
\pgfsetstrokecolor{textcolor}%
\pgfsetfillcolor{textcolor}%
\pgftext[x=0.322915in,y=1.094003in,,bottom,rotate=90.000000]{\color{textcolor}\rmfamily\fontsize{9.000000}{10.800000}\selectfont Voltage [V]}%
\end{pgfscope}%
\begin{pgfscope}%
\pgfpathrectangle{\pgfqpoint{0.639851in}{0.523007in}}{\pgfqpoint{1.625149in}{1.141993in}}%
\pgfusepath{clip}%
\pgfsetrectcap%
\pgfsetroundjoin%
\pgfsetlinewidth{0.702625pt}%
\definecolor{currentstroke}{rgb}{0.250980,0.250980,0.250980}%
\pgfsetstrokecolor{currentstroke}%
\pgfsetdash{}{0pt}%
\pgfpathmoveto{\pgfqpoint{0.639851in}{0.570179in}}%
\pgfpathlineto{\pgfqpoint{0.642112in}{0.587807in}}%
\pgfpathlineto{\pgfqpoint{0.644373in}{0.570179in}}%
\pgfpathlineto{\pgfqpoint{0.649461in}{0.570179in}}%
\pgfpathlineto{\pgfqpoint{0.653983in}{0.587807in}}%
\pgfpathlineto{\pgfqpoint{0.656809in}{0.578993in}}%
\pgfpathlineto{\pgfqpoint{0.659070in}{0.578993in}}%
\pgfpathlineto{\pgfqpoint{0.661331in}{0.587807in}}%
\pgfpathlineto{\pgfqpoint{0.663592in}{0.587807in}}%
\pgfpathlineto{\pgfqpoint{0.666419in}{0.596621in}}%
\pgfpathlineto{\pgfqpoint{0.668680in}{0.578993in}}%
\pgfpathlineto{\pgfqpoint{0.670941in}{0.587807in}}%
\pgfpathlineto{\pgfqpoint{0.673202in}{0.578993in}}%
\pgfpathlineto{\pgfqpoint{0.676028in}{0.587807in}}%
\pgfpathlineto{\pgfqpoint{0.678289in}{0.578993in}}%
\pgfpathlineto{\pgfqpoint{0.680550in}{0.587807in}}%
\pgfpathlineto{\pgfqpoint{0.682812in}{0.587807in}}%
\pgfpathlineto{\pgfqpoint{0.685638in}{0.578993in}}%
\pgfpathlineto{\pgfqpoint{0.687899in}{0.587807in}}%
\pgfpathlineto{\pgfqpoint{0.690160in}{0.570179in}}%
\pgfpathlineto{\pgfqpoint{0.692421in}{0.587807in}}%
\pgfpathlineto{\pgfqpoint{0.695247in}{0.596621in}}%
\pgfpathlineto{\pgfqpoint{0.697509in}{0.587807in}}%
\pgfpathlineto{\pgfqpoint{0.699770in}{0.596621in}}%
\pgfpathlineto{\pgfqpoint{0.704857in}{0.578993in}}%
\pgfpathlineto{\pgfqpoint{0.707118in}{0.587807in}}%
\pgfpathlineto{\pgfqpoint{0.709379in}{0.605435in}}%
\pgfpathlineto{\pgfqpoint{0.711640in}{0.596621in}}%
\pgfpathlineto{\pgfqpoint{0.714467in}{0.596621in}}%
\pgfpathlineto{\pgfqpoint{0.716728in}{0.570179in}}%
\pgfpathlineto{\pgfqpoint{0.718989in}{0.596621in}}%
\pgfpathlineto{\pgfqpoint{0.721250in}{0.587807in}}%
\pgfpathlineto{\pgfqpoint{0.724076in}{0.605435in}}%
\pgfpathlineto{\pgfqpoint{0.726337in}{0.605435in}}%
\pgfpathlineto{\pgfqpoint{0.728598in}{0.596621in}}%
\pgfpathlineto{\pgfqpoint{0.730859in}{0.605435in}}%
\pgfpathlineto{\pgfqpoint{0.745556in}{0.605435in}}%
\pgfpathlineto{\pgfqpoint{0.747817in}{0.623063in}}%
\pgfpathlineto{\pgfqpoint{0.750079in}{0.605435in}}%
\pgfpathlineto{\pgfqpoint{0.752905in}{0.623063in}}%
\pgfpathlineto{\pgfqpoint{0.755166in}{0.605435in}}%
\pgfpathlineto{\pgfqpoint{0.757427in}{0.623063in}}%
\pgfpathlineto{\pgfqpoint{0.759688in}{0.614249in}}%
\pgfpathlineto{\pgfqpoint{0.762514in}{0.623063in}}%
\pgfpathlineto{\pgfqpoint{0.764776in}{0.614249in}}%
\pgfpathlineto{\pgfqpoint{0.767037in}{0.623063in}}%
\pgfpathlineto{\pgfqpoint{0.769298in}{0.640691in}}%
\pgfpathlineto{\pgfqpoint{0.772124in}{0.640691in}}%
\pgfpathlineto{\pgfqpoint{0.774385in}{0.631877in}}%
\pgfpathlineto{\pgfqpoint{0.776646in}{0.640691in}}%
\pgfpathlineto{\pgfqpoint{0.778907in}{0.640691in}}%
\pgfpathlineto{\pgfqpoint{0.781734in}{0.649505in}}%
\pgfpathlineto{\pgfqpoint{0.783995in}{0.640691in}}%
\pgfpathlineto{\pgfqpoint{0.788517in}{0.658319in}}%
\pgfpathlineto{\pgfqpoint{0.791343in}{0.649505in}}%
\pgfpathlineto{\pgfqpoint{0.793604in}{0.658319in}}%
\pgfpathlineto{\pgfqpoint{0.795865in}{0.658319in}}%
\pgfpathlineto{\pgfqpoint{0.798126in}{0.675947in}}%
\pgfpathlineto{\pgfqpoint{0.815085in}{0.675947in}}%
\pgfpathlineto{\pgfqpoint{0.820172in}{0.693575in}}%
\pgfpathlineto{\pgfqpoint{0.822433in}{0.693575in}}%
\pgfpathlineto{\pgfqpoint{0.824694in}{0.684761in}}%
\pgfpathlineto{\pgfqpoint{0.826955in}{0.693575in}}%
\pgfpathlineto{\pgfqpoint{0.832043in}{0.693575in}}%
\pgfpathlineto{\pgfqpoint{0.834304in}{0.684761in}}%
\pgfpathlineto{\pgfqpoint{0.836565in}{0.711203in}}%
\pgfpathlineto{\pgfqpoint{0.839391in}{0.693575in}}%
\pgfpathlineto{\pgfqpoint{0.841652in}{0.711203in}}%
\pgfpathlineto{\pgfqpoint{0.846174in}{0.711203in}}%
\pgfpathlineto{\pgfqpoint{0.849001in}{0.728831in}}%
\pgfpathlineto{\pgfqpoint{0.851262in}{0.720017in}}%
\pgfpathlineto{\pgfqpoint{0.853523in}{0.746459in}}%
\pgfpathlineto{\pgfqpoint{0.860871in}{0.746459in}}%
\pgfpathlineto{\pgfqpoint{0.863132in}{0.764088in}}%
\pgfpathlineto{\pgfqpoint{0.865393in}{0.764088in}}%
\pgfpathlineto{\pgfqpoint{0.868220in}{0.772902in}}%
\pgfpathlineto{\pgfqpoint{0.870481in}{0.772902in}}%
\pgfpathlineto{\pgfqpoint{0.872742in}{0.781716in}}%
\pgfpathlineto{\pgfqpoint{0.877829in}{0.781716in}}%
\pgfpathlineto{\pgfqpoint{0.880090in}{0.799344in}}%
\pgfpathlineto{\pgfqpoint{0.882352in}{0.808158in}}%
\pgfpathlineto{\pgfqpoint{0.884613in}{0.799344in}}%
\pgfpathlineto{\pgfqpoint{0.889700in}{0.816972in}}%
\pgfpathlineto{\pgfqpoint{0.894222in}{0.816972in}}%
\pgfpathlineto{\pgfqpoint{0.897049in}{0.834600in}}%
\pgfpathlineto{\pgfqpoint{0.899310in}{0.825786in}}%
\pgfpathlineto{\pgfqpoint{0.901571in}{0.825786in}}%
\pgfpathlineto{\pgfqpoint{0.906658in}{0.843414in}}%
\pgfpathlineto{\pgfqpoint{0.908919in}{0.834600in}}%
\pgfpathlineto{\pgfqpoint{0.913441in}{0.852228in}}%
\pgfpathlineto{\pgfqpoint{0.916268in}{0.852228in}}%
\pgfpathlineto{\pgfqpoint{0.918529in}{0.869856in}}%
\pgfpathlineto{\pgfqpoint{0.920790in}{0.869856in}}%
\pgfpathlineto{\pgfqpoint{0.923051in}{0.861042in}}%
\pgfpathlineto{\pgfqpoint{0.928138in}{0.887484in}}%
\pgfpathlineto{\pgfqpoint{0.935487in}{0.887484in}}%
\pgfpathlineto{\pgfqpoint{0.937748in}{0.896298in}}%
\pgfpathlineto{\pgfqpoint{0.940009in}{0.887484in}}%
\pgfpathlineto{\pgfqpoint{0.945096in}{0.905112in}}%
\pgfpathlineto{\pgfqpoint{0.949619in}{0.905112in}}%
\pgfpathlineto{\pgfqpoint{0.951880in}{0.922740in}}%
\pgfpathlineto{\pgfqpoint{0.954706in}{0.905112in}}%
\pgfpathlineto{\pgfqpoint{0.956967in}{0.922740in}}%
\pgfpathlineto{\pgfqpoint{0.961489in}{0.922740in}}%
\pgfpathlineto{\pgfqpoint{0.964316in}{0.931554in}}%
\pgfpathlineto{\pgfqpoint{0.966577in}{0.957996in}}%
\pgfpathlineto{\pgfqpoint{0.968838in}{0.931554in}}%
\pgfpathlineto{\pgfqpoint{0.973925in}{0.949182in}}%
\pgfpathlineto{\pgfqpoint{0.976186in}{0.949182in}}%
\pgfpathlineto{\pgfqpoint{0.978447in}{0.940368in}}%
\pgfpathlineto{\pgfqpoint{0.980708in}{0.957996in}}%
\pgfpathlineto{\pgfqpoint{0.983535in}{0.957996in}}%
\pgfpathlineto{\pgfqpoint{0.985796in}{0.966811in}}%
\pgfpathlineto{\pgfqpoint{0.988057in}{0.957996in}}%
\pgfpathlineto{\pgfqpoint{0.990318in}{0.957996in}}%
\pgfpathlineto{\pgfqpoint{0.993144in}{0.975625in}}%
\pgfpathlineto{\pgfqpoint{0.995405in}{0.957996in}}%
\pgfpathlineto{\pgfqpoint{0.999928in}{0.975625in}}%
\pgfpathlineto{\pgfqpoint{1.005015in}{0.975625in}}%
\pgfpathlineto{\pgfqpoint{1.007276in}{0.993253in}}%
\pgfpathlineto{\pgfqpoint{1.009537in}{0.984439in}}%
\pgfpathlineto{\pgfqpoint{1.012363in}{0.984439in}}%
\pgfpathlineto{\pgfqpoint{1.014625in}{0.975625in}}%
\pgfpathlineto{\pgfqpoint{1.016886in}{0.993253in}}%
\pgfpathlineto{\pgfqpoint{1.019147in}{0.984439in}}%
\pgfpathlineto{\pgfqpoint{1.021973in}{0.993253in}}%
\pgfpathlineto{\pgfqpoint{1.024234in}{0.993253in}}%
\pgfpathlineto{\pgfqpoint{1.026495in}{1.002067in}}%
\pgfpathlineto{\pgfqpoint{1.028756in}{0.993253in}}%
\pgfpathlineto{\pgfqpoint{1.031583in}{0.993253in}}%
\pgfpathlineto{\pgfqpoint{1.033844in}{1.010881in}}%
\pgfpathlineto{\pgfqpoint{1.036105in}{1.010881in}}%
\pgfpathlineto{\pgfqpoint{1.041192in}{1.028509in}}%
\pgfpathlineto{\pgfqpoint{1.043453in}{1.028509in}}%
\pgfpathlineto{\pgfqpoint{1.045714in}{1.019695in}}%
\pgfpathlineto{\pgfqpoint{1.050802in}{1.037323in}}%
\pgfpathlineto{\pgfqpoint{1.053063in}{1.037323in}}%
\pgfpathlineto{\pgfqpoint{1.055324in}{1.046137in}}%
\pgfpathlineto{\pgfqpoint{1.057585in}{1.046137in}}%
\pgfpathlineto{\pgfqpoint{1.060411in}{1.063765in}}%
\pgfpathlineto{\pgfqpoint{1.062672in}{1.054951in}}%
\pgfpathlineto{\pgfqpoint{1.064933in}{1.072579in}}%
\pgfpathlineto{\pgfqpoint{1.067195in}{1.072579in}}%
\pgfpathlineto{\pgfqpoint{1.070021in}{1.063765in}}%
\pgfpathlineto{\pgfqpoint{1.072282in}{1.081393in}}%
\pgfpathlineto{\pgfqpoint{1.074543in}{1.081393in}}%
\pgfpathlineto{\pgfqpoint{1.076804in}{1.099021in}}%
\pgfpathlineto{\pgfqpoint{1.079630in}{1.090207in}}%
\pgfpathlineto{\pgfqpoint{1.081892in}{1.099021in}}%
\pgfpathlineto{\pgfqpoint{1.089240in}{1.099021in}}%
\pgfpathlineto{\pgfqpoint{1.091501in}{1.107835in}}%
\pgfpathlineto{\pgfqpoint{1.093762in}{1.099021in}}%
\pgfpathlineto{\pgfqpoint{1.096023in}{1.107835in}}%
\pgfpathlineto{\pgfqpoint{1.098850in}{1.107835in}}%
\pgfpathlineto{\pgfqpoint{1.101111in}{1.116649in}}%
\pgfpathlineto{\pgfqpoint{1.103372in}{1.107835in}}%
\pgfpathlineto{\pgfqpoint{1.105633in}{1.125463in}}%
\pgfpathlineto{\pgfqpoint{1.108459in}{1.116649in}}%
\pgfpathlineto{\pgfqpoint{1.112981in}{1.116649in}}%
\pgfpathlineto{\pgfqpoint{1.115242in}{1.134277in}}%
\pgfpathlineto{\pgfqpoint{1.129939in}{1.134277in}}%
\pgfpathlineto{\pgfqpoint{1.132201in}{1.143091in}}%
\pgfpathlineto{\pgfqpoint{1.134462in}{1.143091in}}%
\pgfpathlineto{\pgfqpoint{1.137288in}{1.151905in}}%
\pgfpathlineto{\pgfqpoint{1.139549in}{1.143091in}}%
\pgfpathlineto{\pgfqpoint{1.141810in}{1.143091in}}%
\pgfpathlineto{\pgfqpoint{1.144071in}{1.134277in}}%
\pgfpathlineto{\pgfqpoint{1.149159in}{1.151905in}}%
\pgfpathlineto{\pgfqpoint{1.151420in}{1.143091in}}%
\pgfpathlineto{\pgfqpoint{1.153681in}{1.160719in}}%
\pgfpathlineto{\pgfqpoint{1.156507in}{1.151905in}}%
\pgfpathlineto{\pgfqpoint{1.161029in}{1.169534in}}%
\pgfpathlineto{\pgfqpoint{1.163290in}{1.169534in}}%
\pgfpathlineto{\pgfqpoint{1.166117in}{1.151905in}}%
\pgfpathlineto{\pgfqpoint{1.168378in}{1.169534in}}%
\pgfpathlineto{\pgfqpoint{1.170639in}{1.160719in}}%
\pgfpathlineto{\pgfqpoint{1.172900in}{1.169534in}}%
\pgfpathlineto{\pgfqpoint{1.180248in}{1.169534in}}%
\pgfpathlineto{\pgfqpoint{1.182509in}{1.187162in}}%
\pgfpathlineto{\pgfqpoint{1.185336in}{1.169534in}}%
\pgfpathlineto{\pgfqpoint{1.189858in}{1.169534in}}%
\pgfpathlineto{\pgfqpoint{1.194945in}{1.187162in}}%
\pgfpathlineto{\pgfqpoint{1.197206in}{1.178348in}}%
\pgfpathlineto{\pgfqpoint{1.199468in}{1.187162in}}%
\pgfpathlineto{\pgfqpoint{1.201729in}{1.187162in}}%
\pgfpathlineto{\pgfqpoint{1.204555in}{1.195976in}}%
\pgfpathlineto{\pgfqpoint{1.206816in}{1.187162in}}%
\pgfpathlineto{\pgfqpoint{1.209077in}{1.195976in}}%
\pgfpathlineto{\pgfqpoint{1.211338in}{1.187162in}}%
\pgfpathlineto{\pgfqpoint{1.214165in}{1.187162in}}%
\pgfpathlineto{\pgfqpoint{1.216426in}{1.204790in}}%
\pgfpathlineto{\pgfqpoint{1.218687in}{1.195976in}}%
\pgfpathlineto{\pgfqpoint{1.220948in}{1.204790in}}%
\pgfpathlineto{\pgfqpoint{1.223774in}{1.195976in}}%
\pgfpathlineto{\pgfqpoint{1.228296in}{1.213604in}}%
\pgfpathlineto{\pgfqpoint{1.230557in}{1.204790in}}%
\pgfpathlineto{\pgfqpoint{1.240167in}{1.231232in}}%
\pgfpathlineto{\pgfqpoint{1.242993in}{1.222418in}}%
\pgfpathlineto{\pgfqpoint{1.245254in}{1.222418in}}%
\pgfpathlineto{\pgfqpoint{1.247515in}{1.240046in}}%
\pgfpathlineto{\pgfqpoint{1.249777in}{1.240046in}}%
\pgfpathlineto{\pgfqpoint{1.252603in}{1.257674in}}%
\pgfpathlineto{\pgfqpoint{1.254864in}{1.240046in}}%
\pgfpathlineto{\pgfqpoint{1.257125in}{1.257674in}}%
\pgfpathlineto{\pgfqpoint{1.259386in}{1.257674in}}%
\pgfpathlineto{\pgfqpoint{1.262212in}{1.266488in}}%
\pgfpathlineto{\pgfqpoint{1.264474in}{1.257674in}}%
\pgfpathlineto{\pgfqpoint{1.266735in}{1.257674in}}%
\pgfpathlineto{\pgfqpoint{1.271822in}{1.275302in}}%
\pgfpathlineto{\pgfqpoint{1.274083in}{1.266488in}}%
\pgfpathlineto{\pgfqpoint{1.276344in}{1.284116in}}%
\pgfpathlineto{\pgfqpoint{1.278605in}{1.275302in}}%
\pgfpathlineto{\pgfqpoint{1.281432in}{1.275302in}}%
\pgfpathlineto{\pgfqpoint{1.283693in}{1.292930in}}%
\pgfpathlineto{\pgfqpoint{1.285954in}{1.284116in}}%
\pgfpathlineto{\pgfqpoint{1.288215in}{1.292930in}}%
\pgfpathlineto{\pgfqpoint{1.291041in}{1.292930in}}%
\pgfpathlineto{\pgfqpoint{1.293302in}{1.275302in}}%
\pgfpathlineto{\pgfqpoint{1.295563in}{1.292930in}}%
\pgfpathlineto{\pgfqpoint{1.297824in}{1.292930in}}%
\pgfpathlineto{\pgfqpoint{1.300651in}{1.301744in}}%
\pgfpathlineto{\pgfqpoint{1.302912in}{1.301744in}}%
\pgfpathlineto{\pgfqpoint{1.305173in}{1.310558in}}%
\pgfpathlineto{\pgfqpoint{1.314782in}{1.310558in}}%
\pgfpathlineto{\pgfqpoint{1.319870in}{1.328186in}}%
\pgfpathlineto{\pgfqpoint{1.322131in}{1.319372in}}%
\pgfpathlineto{\pgfqpoint{1.324392in}{1.328186in}}%
\pgfpathlineto{\pgfqpoint{1.336263in}{1.328186in}}%
\pgfpathlineto{\pgfqpoint{1.339089in}{1.337000in}}%
\pgfpathlineto{\pgfqpoint{1.341350in}{1.337000in}}%
\pgfpathlineto{\pgfqpoint{1.343611in}{1.345814in}}%
\pgfpathlineto{\pgfqpoint{1.345872in}{1.337000in}}%
\pgfpathlineto{\pgfqpoint{1.350960in}{1.337000in}}%
\pgfpathlineto{\pgfqpoint{1.353221in}{1.345814in}}%
\pgfpathlineto{\pgfqpoint{1.382050in}{1.345814in}}%
\pgfpathlineto{\pgfqpoint{1.384311in}{1.354628in}}%
\pgfpathlineto{\pgfqpoint{1.387137in}{1.354628in}}%
\pgfpathlineto{\pgfqpoint{1.389398in}{1.363442in}}%
\pgfpathlineto{\pgfqpoint{1.393920in}{1.363442in}}%
\pgfpathlineto{\pgfqpoint{1.396747in}{1.372256in}}%
\pgfpathlineto{\pgfqpoint{1.399008in}{1.363442in}}%
\pgfpathlineto{\pgfqpoint{1.410878in}{1.363442in}}%
\pgfpathlineto{\pgfqpoint{1.413139in}{1.372256in}}%
\pgfpathlineto{\pgfqpoint{1.415966in}{1.372256in}}%
\pgfpathlineto{\pgfqpoint{1.418227in}{1.381071in}}%
\pgfpathlineto{\pgfqpoint{1.420488in}{1.363442in}}%
\pgfpathlineto{\pgfqpoint{1.422749in}{1.381071in}}%
\pgfpathlineto{\pgfqpoint{1.427836in}{1.381071in}}%
\pgfpathlineto{\pgfqpoint{1.430097in}{1.389885in}}%
\pgfpathlineto{\pgfqpoint{1.432358in}{1.381071in}}%
\pgfpathlineto{\pgfqpoint{1.435185in}{1.389885in}}%
\pgfpathlineto{\pgfqpoint{1.437446in}{1.416327in}}%
\pgfpathlineto{\pgfqpoint{1.439707in}{1.398699in}}%
\pgfpathlineto{\pgfqpoint{1.441968in}{1.416327in}}%
\pgfpathlineto{\pgfqpoint{1.451578in}{1.416327in}}%
\pgfpathlineto{\pgfqpoint{1.454404in}{1.425141in}}%
\pgfpathlineto{\pgfqpoint{1.458926in}{1.425141in}}%
\pgfpathlineto{\pgfqpoint{1.461187in}{1.433955in}}%
\pgfpathlineto{\pgfqpoint{1.468536in}{1.433955in}}%
\pgfpathlineto{\pgfqpoint{1.470797in}{1.442769in}}%
\pgfpathlineto{\pgfqpoint{1.473623in}{1.442769in}}%
\pgfpathlineto{\pgfqpoint{1.475884in}{1.451583in}}%
\pgfpathlineto{\pgfqpoint{1.480406in}{1.451583in}}%
\pgfpathlineto{\pgfqpoint{1.483233in}{1.433955in}}%
\pgfpathlineto{\pgfqpoint{1.485494in}{1.451583in}}%
\pgfpathlineto{\pgfqpoint{1.490016in}{1.451583in}}%
\pgfpathlineto{\pgfqpoint{1.492842in}{1.442769in}}%
\pgfpathlineto{\pgfqpoint{1.495103in}{1.451583in}}%
\pgfpathlineto{\pgfqpoint{1.499626in}{1.451583in}}%
\pgfpathlineto{\pgfqpoint{1.502452in}{1.442769in}}%
\pgfpathlineto{\pgfqpoint{1.504713in}{1.469211in}}%
\pgfpathlineto{\pgfqpoint{1.506974in}{1.451583in}}%
\pgfpathlineto{\pgfqpoint{1.533542in}{1.451583in}}%
\pgfpathlineto{\pgfqpoint{1.535803in}{1.442769in}}%
\pgfpathlineto{\pgfqpoint{1.538064in}{1.442769in}}%
\pgfpathlineto{\pgfqpoint{1.540890in}{1.451583in}}%
\pgfpathlineto{\pgfqpoint{1.543151in}{1.442769in}}%
\pgfpathlineto{\pgfqpoint{1.545412in}{1.451583in}}%
\pgfpathlineto{\pgfqpoint{1.547673in}{1.451583in}}%
\pgfpathlineto{\pgfqpoint{1.550500in}{1.442769in}}%
\pgfpathlineto{\pgfqpoint{1.552761in}{1.451583in}}%
\pgfpathlineto{\pgfqpoint{1.555022in}{1.433955in}}%
\pgfpathlineto{\pgfqpoint{1.562370in}{1.433955in}}%
\pgfpathlineto{\pgfqpoint{1.564631in}{1.451583in}}%
\pgfpathlineto{\pgfqpoint{1.566893in}{1.451583in}}%
\pgfpathlineto{\pgfqpoint{1.569719in}{1.425141in}}%
\pgfpathlineto{\pgfqpoint{1.571980in}{1.425141in}}%
\pgfpathlineto{\pgfqpoint{1.574241in}{1.416327in}}%
\pgfpathlineto{\pgfqpoint{1.576502in}{1.433955in}}%
\pgfpathlineto{\pgfqpoint{1.579328in}{1.416327in}}%
\pgfpathlineto{\pgfqpoint{1.581590in}{1.416327in}}%
\pgfpathlineto{\pgfqpoint{1.583851in}{1.433955in}}%
\pgfpathlineto{\pgfqpoint{1.586112in}{1.416327in}}%
\pgfpathlineto{\pgfqpoint{1.598548in}{1.416327in}}%
\pgfpathlineto{\pgfqpoint{1.600809in}{1.389885in}}%
\pgfpathlineto{\pgfqpoint{1.603070in}{1.407513in}}%
\pgfpathlineto{\pgfqpoint{1.605331in}{1.398699in}}%
\pgfpathlineto{\pgfqpoint{1.612679in}{1.398699in}}%
\pgfpathlineto{\pgfqpoint{1.614940in}{1.389885in}}%
\pgfpathlineto{\pgfqpoint{1.620028in}{1.389885in}}%
\pgfpathlineto{\pgfqpoint{1.622289in}{1.398699in}}%
\pgfpathlineto{\pgfqpoint{1.624550in}{1.381071in}}%
\pgfpathlineto{\pgfqpoint{1.627376in}{1.372256in}}%
\pgfpathlineto{\pgfqpoint{1.629637in}{1.381071in}}%
\pgfpathlineto{\pgfqpoint{1.631898in}{1.381071in}}%
\pgfpathlineto{\pgfqpoint{1.634160in}{1.372256in}}%
\pgfpathlineto{\pgfqpoint{1.636986in}{1.381071in}}%
\pgfpathlineto{\pgfqpoint{1.639247in}{1.354628in}}%
\pgfpathlineto{\pgfqpoint{1.641508in}{1.363442in}}%
\pgfpathlineto{\pgfqpoint{1.651118in}{1.363442in}}%
\pgfpathlineto{\pgfqpoint{1.653379in}{1.345814in}}%
\pgfpathlineto{\pgfqpoint{1.665815in}{1.345814in}}%
\pgfpathlineto{\pgfqpoint{1.670337in}{1.328186in}}%
\pgfpathlineto{\pgfqpoint{1.672598in}{1.328186in}}%
\pgfpathlineto{\pgfqpoint{1.675424in}{1.345814in}}%
\pgfpathlineto{\pgfqpoint{1.677685in}{1.328186in}}%
\pgfpathlineto{\pgfqpoint{1.685034in}{1.328186in}}%
\pgfpathlineto{\pgfqpoint{1.687295in}{1.310558in}}%
\pgfpathlineto{\pgfqpoint{1.689556in}{1.328186in}}%
\pgfpathlineto{\pgfqpoint{1.694643in}{1.310558in}}%
\pgfpathlineto{\pgfqpoint{1.696904in}{1.319372in}}%
\pgfpathlineto{\pgfqpoint{1.699166in}{1.301744in}}%
\pgfpathlineto{\pgfqpoint{1.701427in}{1.310558in}}%
\pgfpathlineto{\pgfqpoint{1.704253in}{1.301744in}}%
\pgfpathlineto{\pgfqpoint{1.708775in}{1.301744in}}%
\pgfpathlineto{\pgfqpoint{1.711036in}{1.310558in}}%
\pgfpathlineto{\pgfqpoint{1.713863in}{1.310558in}}%
\pgfpathlineto{\pgfqpoint{1.718385in}{1.292930in}}%
\pgfpathlineto{\pgfqpoint{1.720646in}{1.301744in}}%
\pgfpathlineto{\pgfqpoint{1.723472in}{1.292930in}}%
\pgfpathlineto{\pgfqpoint{1.725733in}{1.275302in}}%
\pgfpathlineto{\pgfqpoint{1.727994in}{1.275302in}}%
\pgfpathlineto{\pgfqpoint{1.730255in}{1.292930in}}%
\pgfpathlineto{\pgfqpoint{1.733082in}{1.275302in}}%
\pgfpathlineto{\pgfqpoint{1.735343in}{1.275302in}}%
\pgfpathlineto{\pgfqpoint{1.737604in}{1.257674in}}%
\pgfpathlineto{\pgfqpoint{1.739865in}{1.275302in}}%
\pgfpathlineto{\pgfqpoint{1.742691in}{1.275302in}}%
\pgfpathlineto{\pgfqpoint{1.744952in}{1.257674in}}%
\pgfpathlineto{\pgfqpoint{1.749474in}{1.257674in}}%
\pgfpathlineto{\pgfqpoint{1.752301in}{1.266488in}}%
\pgfpathlineto{\pgfqpoint{1.754562in}{1.240046in}}%
\pgfpathlineto{\pgfqpoint{1.756823in}{1.257674in}}%
\pgfpathlineto{\pgfqpoint{1.759084in}{1.257674in}}%
\pgfpathlineto{\pgfqpoint{1.761910in}{1.240046in}}%
\pgfpathlineto{\pgfqpoint{1.764171in}{1.248860in}}%
\pgfpathlineto{\pgfqpoint{1.766433in}{1.248860in}}%
\pgfpathlineto{\pgfqpoint{1.768694in}{1.240046in}}%
\pgfpathlineto{\pgfqpoint{1.776042in}{1.240046in}}%
\pgfpathlineto{\pgfqpoint{1.778303in}{1.222418in}}%
\pgfpathlineto{\pgfqpoint{1.785652in}{1.222418in}}%
\pgfpathlineto{\pgfqpoint{1.787913in}{1.213604in}}%
\pgfpathlineto{\pgfqpoint{1.790739in}{1.213604in}}%
\pgfpathlineto{\pgfqpoint{1.793000in}{1.222418in}}%
\pgfpathlineto{\pgfqpoint{1.797522in}{1.204790in}}%
\pgfpathlineto{\pgfqpoint{1.807132in}{1.204790in}}%
\pgfpathlineto{\pgfqpoint{1.809958in}{1.195976in}}%
\pgfpathlineto{\pgfqpoint{1.812219in}{1.204790in}}%
\pgfpathlineto{\pgfqpoint{1.814480in}{1.187162in}}%
\pgfpathlineto{\pgfqpoint{1.821829in}{1.187162in}}%
\pgfpathlineto{\pgfqpoint{1.824090in}{1.178348in}}%
\pgfpathlineto{\pgfqpoint{1.826351in}{1.187162in}}%
\pgfpathlineto{\pgfqpoint{1.829177in}{1.187162in}}%
\pgfpathlineto{\pgfqpoint{1.833700in}{1.169534in}}%
\pgfpathlineto{\pgfqpoint{1.838787in}{1.169534in}}%
\pgfpathlineto{\pgfqpoint{1.841048in}{1.178348in}}%
\pgfpathlineto{\pgfqpoint{1.843309in}{1.169534in}}%
\pgfpathlineto{\pgfqpoint{1.845570in}{1.169534in}}%
\pgfpathlineto{\pgfqpoint{1.848397in}{1.151905in}}%
\pgfpathlineto{\pgfqpoint{1.852919in}{1.169534in}}%
\pgfpathlineto{\pgfqpoint{1.855180in}{1.151905in}}%
\pgfpathlineto{\pgfqpoint{1.858006in}{1.160719in}}%
\pgfpathlineto{\pgfqpoint{1.860267in}{1.143091in}}%
\pgfpathlineto{\pgfqpoint{1.862528in}{1.151905in}}%
\pgfpathlineto{\pgfqpoint{1.867616in}{1.134277in}}%
\pgfpathlineto{\pgfqpoint{1.877225in}{1.134277in}}%
\pgfpathlineto{\pgfqpoint{1.879486in}{1.125463in}}%
\pgfpathlineto{\pgfqpoint{1.881747in}{1.134277in}}%
\pgfpathlineto{\pgfqpoint{1.886835in}{1.116649in}}%
\pgfpathlineto{\pgfqpoint{1.889096in}{1.116649in}}%
\pgfpathlineto{\pgfqpoint{1.891357in}{1.125463in}}%
\pgfpathlineto{\pgfqpoint{1.896444in}{1.107835in}}%
\pgfpathlineto{\pgfqpoint{1.900967in}{1.107835in}}%
\pgfpathlineto{\pgfqpoint{1.903228in}{1.099021in}}%
\pgfpathlineto{\pgfqpoint{1.906054in}{1.107835in}}%
\pgfpathlineto{\pgfqpoint{1.908315in}{1.099021in}}%
\pgfpathlineto{\pgfqpoint{1.915664in}{1.099021in}}%
\pgfpathlineto{\pgfqpoint{1.917925in}{1.081393in}}%
\pgfpathlineto{\pgfqpoint{1.934883in}{1.081393in}}%
\pgfpathlineto{\pgfqpoint{1.937144in}{1.063765in}}%
\pgfpathlineto{\pgfqpoint{1.939405in}{1.081393in}}%
\pgfpathlineto{\pgfqpoint{1.941666in}{1.063765in}}%
\pgfpathlineto{\pgfqpoint{1.944492in}{1.063765in}}%
\pgfpathlineto{\pgfqpoint{1.946753in}{1.054951in}}%
\pgfpathlineto{\pgfqpoint{1.949015in}{1.063765in}}%
\pgfpathlineto{\pgfqpoint{1.956363in}{1.063765in}}%
\pgfpathlineto{\pgfqpoint{1.958624in}{1.054951in}}%
\pgfpathlineto{\pgfqpoint{1.960885in}{1.063765in}}%
\pgfpathlineto{\pgfqpoint{1.963712in}{1.063765in}}%
\pgfpathlineto{\pgfqpoint{1.965973in}{1.046137in}}%
\pgfpathlineto{\pgfqpoint{1.968234in}{1.046137in}}%
\pgfpathlineto{\pgfqpoint{1.970495in}{1.037323in}}%
\pgfpathlineto{\pgfqpoint{1.973321in}{1.046137in}}%
\pgfpathlineto{\pgfqpoint{1.977843in}{1.028509in}}%
\pgfpathlineto{\pgfqpoint{1.982931in}{1.046137in}}%
\pgfpathlineto{\pgfqpoint{1.985192in}{1.028509in}}%
\pgfpathlineto{\pgfqpoint{1.992540in}{1.028509in}}%
\pgfpathlineto{\pgfqpoint{1.994801in}{1.037323in}}%
\pgfpathlineto{\pgfqpoint{1.997062in}{1.028509in}}%
\pgfpathlineto{\pgfqpoint{2.008933in}{1.028509in}}%
\pgfpathlineto{\pgfqpoint{2.011759in}{1.010881in}}%
\pgfpathlineto{\pgfqpoint{2.023630in}{1.010881in}}%
\pgfpathlineto{\pgfqpoint{2.025891in}{1.002067in}}%
\pgfpathlineto{\pgfqpoint{2.028152in}{1.010881in}}%
\pgfpathlineto{\pgfqpoint{2.030979in}{0.993253in}}%
\pgfpathlineto{\pgfqpoint{2.033240in}{1.002067in}}%
\pgfpathlineto{\pgfqpoint{2.035501in}{0.993253in}}%
\pgfpathlineto{\pgfqpoint{2.047371in}{0.993253in}}%
\pgfpathlineto{\pgfqpoint{2.050198in}{0.975625in}}%
\pgfpathlineto{\pgfqpoint{2.052459in}{0.975625in}}%
\pgfpathlineto{\pgfqpoint{2.054720in}{0.984439in}}%
\pgfpathlineto{\pgfqpoint{2.056981in}{0.975625in}}%
\pgfpathlineto{\pgfqpoint{2.059807in}{0.984439in}}%
\pgfpathlineto{\pgfqpoint{2.062068in}{0.966811in}}%
\pgfpathlineto{\pgfqpoint{2.064329in}{0.975625in}}%
\pgfpathlineto{\pgfqpoint{2.066591in}{0.966811in}}%
\pgfpathlineto{\pgfqpoint{2.069417in}{0.966811in}}%
\pgfpathlineto{\pgfqpoint{2.071678in}{0.975625in}}%
\pgfpathlineto{\pgfqpoint{2.073939in}{0.975625in}}%
\pgfpathlineto{\pgfqpoint{2.079026in}{0.957996in}}%
\pgfpathlineto{\pgfqpoint{2.081288in}{0.966811in}}%
\pgfpathlineto{\pgfqpoint{2.083549in}{0.966811in}}%
\pgfpathlineto{\pgfqpoint{2.085810in}{0.957996in}}%
\pgfpathlineto{\pgfqpoint{2.098246in}{0.957996in}}%
\pgfpathlineto{\pgfqpoint{2.100507in}{0.940368in}}%
\pgfpathlineto{\pgfqpoint{2.102768in}{0.957996in}}%
\pgfpathlineto{\pgfqpoint{2.105029in}{0.940368in}}%
\pgfpathlineto{\pgfqpoint{2.107855in}{0.949182in}}%
\pgfpathlineto{\pgfqpoint{2.110116in}{0.940368in}}%
\pgfpathlineto{\pgfqpoint{2.112377in}{0.940368in}}%
\pgfpathlineto{\pgfqpoint{2.114638in}{0.931554in}}%
\pgfpathlineto{\pgfqpoint{2.117465in}{0.931554in}}%
\pgfpathlineto{\pgfqpoint{2.119726in}{0.940368in}}%
\pgfpathlineto{\pgfqpoint{2.121987in}{0.931554in}}%
\pgfpathlineto{\pgfqpoint{2.124248in}{0.931554in}}%
\pgfpathlineto{\pgfqpoint{2.127074in}{0.922740in}}%
\pgfpathlineto{\pgfqpoint{2.131596in}{0.940368in}}%
\pgfpathlineto{\pgfqpoint{2.136684in}{0.922740in}}%
\pgfpathlineto{\pgfqpoint{2.141206in}{0.922740in}}%
\pgfpathlineto{\pgfqpoint{2.143467in}{0.940368in}}%
\pgfpathlineto{\pgfqpoint{2.146293in}{0.922740in}}%
\pgfpathlineto{\pgfqpoint{2.155903in}{0.922740in}}%
\pgfpathlineto{\pgfqpoint{2.158164in}{0.940368in}}%
\pgfpathlineto{\pgfqpoint{2.160425in}{0.922740in}}%
\pgfpathlineto{\pgfqpoint{2.162686in}{0.922740in}}%
\pgfpathlineto{\pgfqpoint{2.165513in}{0.913926in}}%
\pgfpathlineto{\pgfqpoint{2.167774in}{0.922740in}}%
\pgfpathlineto{\pgfqpoint{2.170035in}{0.905112in}}%
\pgfpathlineto{\pgfqpoint{2.172296in}{0.913926in}}%
\pgfpathlineto{\pgfqpoint{2.175122in}{0.905112in}}%
\pgfpathlineto{\pgfqpoint{2.184732in}{0.905112in}}%
\pgfpathlineto{\pgfqpoint{2.186993in}{0.887484in}}%
\pgfpathlineto{\pgfqpoint{2.189254in}{0.905112in}}%
\pgfpathlineto{\pgfqpoint{2.191515in}{0.887484in}}%
\pgfpathlineto{\pgfqpoint{2.194341in}{0.896298in}}%
\pgfpathlineto{\pgfqpoint{2.196602in}{0.887484in}}%
\pgfpathlineto{\pgfqpoint{2.225431in}{0.887484in}}%
\pgfpathlineto{\pgfqpoint{2.229953in}{0.869856in}}%
\pgfpathlineto{\pgfqpoint{2.235041in}{0.887484in}}%
\pgfpathlineto{\pgfqpoint{2.237302in}{0.869856in}}%
\pgfpathlineto{\pgfqpoint{2.249172in}{0.869856in}}%
\pgfpathlineto{\pgfqpoint{2.251999in}{0.852228in}}%
\pgfpathlineto{\pgfqpoint{2.258782in}{0.852228in}}%
\pgfpathlineto{\pgfqpoint{2.261608in}{0.861042in}}%
\pgfpathlineto{\pgfqpoint{2.263869in}{0.852228in}}%
\pgfpathlineto{\pgfqpoint{2.266131in}{0.852228in}}%
\pgfpathlineto{\pgfqpoint{2.266131in}{0.852228in}}%
\pgfusepath{stroke}%
\end{pgfscope}%
\begin{pgfscope}%
\pgfpathrectangle{\pgfqpoint{0.639851in}{0.523007in}}{\pgfqpoint{1.625149in}{1.141993in}}%
\pgfusepath{clip}%
\pgfsetbuttcap%
\pgfsetroundjoin%
\pgfsetlinewidth{0.501875pt}%
\definecolor{currentstroke}{rgb}{0.250980,0.250980,0.250980}%
\pgfsetstrokecolor{currentstroke}%
\pgfsetdash{{1.850000pt}{0.800000pt}}{0.000000pt}%
\pgfpathmoveto{\pgfqpoint{1.540749in}{0.523007in}}%
\pgfpathlineto{\pgfqpoint{1.540749in}{1.493701in}}%
\pgfusepath{stroke}%
\end{pgfscope}%
\begin{pgfscope}%
\pgfsetrectcap%
\pgfsetmiterjoin%
\pgfsetlinewidth{0.803000pt}%
\definecolor{currentstroke}{rgb}{0.000000,0.000000,0.000000}%
\pgfsetstrokecolor{currentstroke}%
\pgfsetdash{}{0pt}%
\pgfpathmoveto{\pgfqpoint{0.639851in}{0.523007in}}%
\pgfpathlineto{\pgfqpoint{0.639851in}{1.665000in}}%
\pgfusepath{stroke}%
\end{pgfscope}%
\begin{pgfscope}%
\pgfsetrectcap%
\pgfsetmiterjoin%
\pgfsetlinewidth{0.803000pt}%
\definecolor{currentstroke}{rgb}{0.000000,0.000000,0.000000}%
\pgfsetstrokecolor{currentstroke}%
\pgfsetdash{}{0pt}%
\pgfpathmoveto{\pgfqpoint{2.265000in}{0.523007in}}%
\pgfpathlineto{\pgfqpoint{2.265000in}{1.665000in}}%
\pgfusepath{stroke}%
\end{pgfscope}%
\begin{pgfscope}%
\pgfsetrectcap%
\pgfsetmiterjoin%
\pgfsetlinewidth{0.803000pt}%
\definecolor{currentstroke}{rgb}{0.000000,0.000000,0.000000}%
\pgfsetstrokecolor{currentstroke}%
\pgfsetdash{}{0pt}%
\pgfpathmoveto{\pgfqpoint{0.639851in}{0.523007in}}%
\pgfpathlineto{\pgfqpoint{2.265000in}{0.523007in}}%
\pgfusepath{stroke}%
\end{pgfscope}%
\begin{pgfscope}%
\pgfsetrectcap%
\pgfsetmiterjoin%
\pgfsetlinewidth{0.803000pt}%
\definecolor{currentstroke}{rgb}{0.000000,0.000000,0.000000}%
\pgfsetstrokecolor{currentstroke}%
\pgfsetdash{}{0pt}%
\pgfpathmoveto{\pgfqpoint{0.639851in}{1.665000in}}%
\pgfpathlineto{\pgfqpoint{2.265000in}{1.665000in}}%
\pgfusepath{stroke}%
\end{pgfscope}%
\begin{pgfscope}%
\definecolor{textcolor}{rgb}{0.000000,0.000000,0.000000}%
\pgfsetstrokecolor{textcolor}%
\pgfsetfillcolor{textcolor}%
\pgftext[x=1.090300in,y=1.501858in,,bottom]{\color{textcolor}\rmfamily\fontsize{8.000000}{9.600000}\selectfont integration}%
\end{pgfscope}%
\begin{pgfscope}%
\definecolor{textcolor}{rgb}{0.000000,0.000000,0.000000}%
\pgfsetstrokecolor{textcolor}%
\pgfsetfillcolor{textcolor}%
\pgftext[x=1.902874in,y=1.501858in,,bottom]{\color{textcolor}\rmfamily\fontsize{8.000000}{9.600000}\selectfont decay}%
\end{pgfscope}%
\end{pgfpicture}%
\makeatother%
\endgroup%

%% file: tex/flowchart_mac.tikz
\tikzset{%
  >={Latex[width=2mm,length=2mm]},
            base/.style = {rectangle, rounded corners, draw=black,
                           minimum width=3cm, minimum height=0.6cm,
                           text centered, font=\rmfamily},
}

\begin{tikzpicture}[node distance=4cm,
    every node/.style={fill=white, font=\sffamily}, align=center]
  \node (write_synapse_matrix)[base] {set up matrix};
  \node (reset_neurons) [base, below of=write_synapse_matrix, yshift=3cm] {reset neurons};
  \node (send_inputs) [base, right of=reset_neurons] {send inputs};
  \node (read_activations) [base, right of=send_inputs] {read activations};
 
  \draw[->]  (write_synapse_matrix) -- (reset_neurons);
  \draw[->]         (reset_neurons) -- (send_inputs);
  \draw[->]           (send_inputs) -- (read_activations);
  \draw[->]  (read_activations.north) -- ++(0,0.7) -- ++(-5.5,0) -- 
    node[xshift=6cm, yshift=0.35cm]{\rmfamily next vector} (reset_neurons.north east);
  \path[->]  (send_inputs) edge [gray, loop above, distance=0.8cm]
    node[xshift=0.85cm,yshift=-0.45cm]{\rmfamily resends} (send_inputs);
\end{tikzpicture}

%% file: tex/3_results.tex
\section{Results}\label{sec:results}

\subsection{Characterization}

\begin{figure}[t]
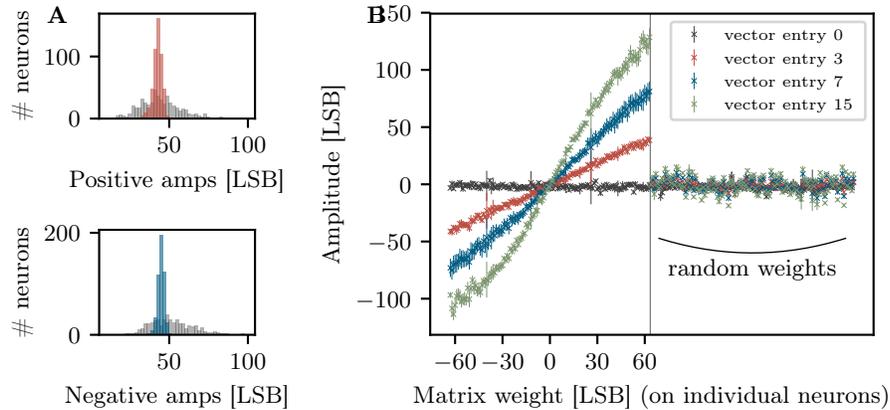

    \begin{subfigure}{0.33\textwidth}
        \begin{tikzpicture}
            \draw (0, 0) node[anchor=north west, inner sep=0] {
                \input{plots/neuron_histograms.pgf}
            };
            \node[inner sep=0pt] at (1., -0.45) {\textbf{A}};
        \end{tikzpicture}
    \end{subfigure}
    \hfill
    \begin{subfigure}{0.66\textwidth}
        \begin{tikzpicture}
            \draw (0, 0) node[anchor=north west, inner sep=0] {
                \input{plots/multiplication_top.pgf}
            };
            \node[inner sep=0pt] at (1.1, -0.45) {\textbf{B}};
        \end{tikzpicture}
    \end{subfigure}
    \caption{
        \textbf{A}:
        Histogram of amplitudes received on all neurons for equal inputs.
        After calibration (colored) the width of the distribution is decreased compared to the uncalibrated state (gray).
        \textbf{B}:
        Characterization of analog multiplication results, sweeping both matrix weights and vector entries.
        The left half of the synapse matrix was configured such that weights increased from a value of -63 in the leftmost column to +63 in the 127th column with an increment of one.
        Within each column, all weights were set to the same value.
        The right half was set to random weights for each synapse.
        We injected four constant input vectors, each consisting of 128 entries of 0, 3, 7 and 15.
        Error bars indicate the standard deviation within 30 runs.
        \todo[inline]{ECM: is plotting of standard deviation valid? how does min/max look like (``schlauchplot'')?}
    }
    \label{fig:characterization}
\end{figure}

We first evaluated the performance of \BrainScaleS{2}'s matrix multiplication mode by configuring a synthetic test matrix.
In the left half, weights increased linearly from left to right, all synapses in column $i$ were set to weight $w = i - 63$.
In the right half, each synapse was set to a random weight, drawn uniformly from -63 to 63.
Multiple homogeneous vectors of different amplitude were used to characterize the linearity of both vector entries and matrix weights (\cref{fig:characterization}B).

For lower weights and inputs, the multiplication followed the expected linear behavior.
For higher activations, saturation occurred, which is most clearly observable for the vector with entries of 15.
However, we expect most real-world networks to use sparser matrices with more balanced excitatory and inhibitory weights than this test.
In the right part of the matrix, the random weights resulted in activations close to zero.
It is notable that per column this activation was either positive or negative, depending on the exact mean of the associated weights and scaling with the injected vectors' values.
This suggests that both, positive and nagative inputs can be summed correctly and have been tuned to the same strength.

\subsection{MNIST benchmark}

The above measurements indicate that the analog substrate can indeed be used to perform MAC operations.
To investigate its performance on a common benchmark, the MNIST \citep{lecunmnist} dataset is used.
\citet{spilger2020hxtorch} additionally trained and classified the Human Activity Recognition dataset on \BrainScaleS{2}.

\subsubsection{Models}
We examined the performance of two different network models.
Both relied on ReLU activation functions for the hidden layers and a softmax for the output layer.
The two networks did not incorporate a bias.
They were trained with the Adam optimizer \citep{kingma2014adam} in TensorFlow \citep{tensorflow2015} using 32 bit float weights.

The \emph{convolutional model}
was based on images zero-padded by one to \num{30}\texttimes{}\num{30} pixels.
It consisted of a two-dimensional convolutional layer, a dense layer of \num{128} neurons and 10 label units.
The convolution layer used \num{20} \num{10}\texttimes{}\num{10} filters with a stride of \num{5}\texttimes{}\num{5}.

The \emph{dense model} consisted of two fully connected layers of \num{64} and ten neurons, respectively.
Due to its small size, this model directly fits into the on-chip weight matrix and does not require reconfiguration.

\subsubsection{Accuracy}

As a first step in transferring the models to \BrainScaleS{2}, we discretized the weights to 6 bit integers plus sign.
Evaluating both networks in software showed only slight drops in performance (\cref{table:mnist_results}).

\begin{figure}[t]
    \begin{center}
        \input{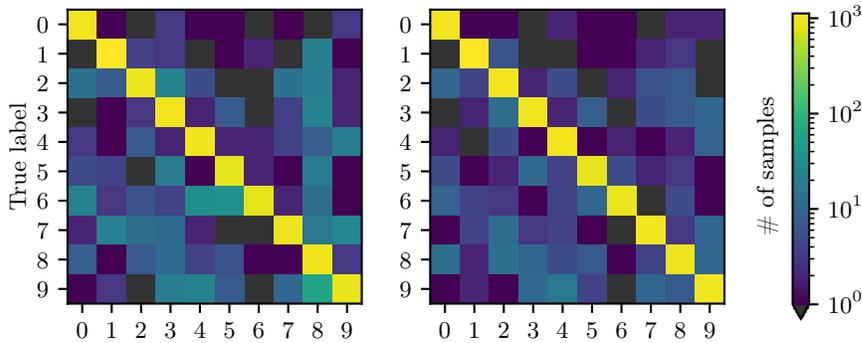}
    \end{center}
    \caption{
        Confusion matrix of the dense network running MNIST.
        Left: Executed pre-trained model on hardware, no re-training.
        Right: Results after training one epoch with hardware in the loop.
        Note the logarithmic colorbar in both plots.
    }
    \label{fig:mnist_confusion}
\end{figure}

When executing the convolutional network on \BrainScaleS{2}, the classification performance on held-out test data dropped  to \SI{92.1}{\%}, indicating a non-ideal calibration of and temporal noise in the analog circuits.
After a continued training with hardware in the loop, an accuracy of \SI{98.0}{\%} could be restored, closely matching the results obtained in software.
A similar behavior was observed for the smaller dense model, with \SI{96.3}{\percent} after training in the loop.
Here, the discrepancy between software and hardware performance was more pronounced, we suspect that with the smaller number of synapses involved, the potential for correction is also lower as there is less redundancy.
For the dense network, confusion matrices are shown in~\cref{fig:mnist_confusion}.
Classification works for all digits, no systematic misclassification is observable.

\begin{table}[t]
    \centering
    \caption{
        MNIST classification accuracy in percent for two models in different conditions.
        The networks were trained in software using 32 bit float weights.
        Discretization to 6 bit integer weights plus sign had little impact on the performance.
        Transferring the network to the chip lead to a loss of accuracy, which could be be restored after training with hardware in the loop.
    }
    \vspace{0.5em}
    \begin{tabular}{l@{\hskip.25cm}c@{\hskip.25cm}c@{\hskip.25cm}c@{\hskip.25cm}c}
        & \multicolumn{2}{c}{software} & \multicolumn{2}{c}{hardware} \\
        & 32 bit & 6 bit & calibration & trained \\
        & float  & int   & only        & in the loop \\
        \toprule
        convolutional model & 98.29\,\% & 98.10\,\% & 92.13\,\% & 98.01\,\% \\
        dense model & 97.43\,\% & 97.36\,\% & 92.46\,\% & 96.30\,\% \\
    \end{tabular}
    \label{table:mnist_results}
\end{table}

\begin{figure}[t]
    \begin{center}
        \input{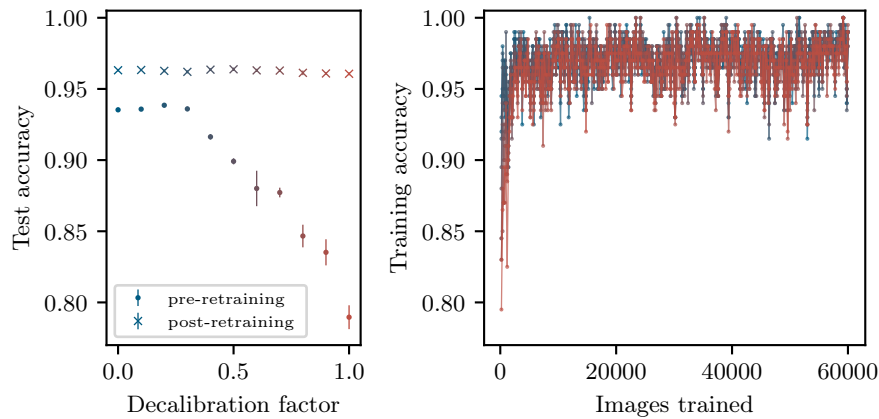}
    \end{center}
    \caption{
        MNIST accuracy when detuning the calibrated neuron parameters, shown for the dense network.
        Left: Accuracy before and after training one epoch with hardware in the loop.
        Results show the mean and standard deviation of 10 runs classifying the \num{10000} test images with unchanged parameters.
        Right: Accuracy per batch during the one epoch of training.
        \num{200} images per batch, \num{300} batches per epoch.
        Colors indicate the state of calibration, corresponding to the left plot.
        \todoforJWW[inline]{SB/YS: Replace training accuracy in right plot by test accuracy. (JW: This takes months (?) to measure...)}
    }
    \label{fig:mnist_decalibration}
\end{figure}

\subsubsection{Energy consumption}
\label{sec:energy-consumption}

A bug in the current chip revision requires constant reconfiguration of the synapse matrix for each input vector.
It yields a drastically increased runtime and energy consumption, far off the targeted performance for \BrainScaleS{2}.
We expect the performance to increase by two to three orders of magnitude for the next chip revision.

Currently, when fully utilizing the resources on hardware, vectors of up to 256 entries can be multiplied with matrices of up to 512 columns.
One of these vector-matrix multiplications takes \SI{5}{\milli\second}, and at a power consumption of approximately \SI{0.3}{\watt} \citep{cramer2020training}, takes \SI{1.5}{\milli\joule}.
In signed mode, four multiplications involving only a subset of neurons can be executed simultaneously, all receiving independent full-size vector inputs.
This is possible utilizing the synapse label bits and both synapse matrices on chip.

Assuming parallelized and batched execution, inference with the convolutional network takes \SI{40}{\milli\second} per image, resulting in an energy consumption of \SI{12}{\milli\joule}.
The smaller dense network, again assuming parallel execution, requires \SI{10}{\milli\second} and therefore \SI{3}{\milli\joule} per image.

\subsubsection{Calibration vs. learning}

It has previously been shown that, to a certain degree, learning can replace the need for explicit calibration \citep{wunderlich2019demonstrating}.
To analyze such effects for the presented models, we started with a software-trained network and then detuned the calibration of the neurons' leak conductance and synaptic input amplitudes by continuously transitioning between the calibrated and uncalibrated state.
The latter resulted from taking the median of the respective parameter distribution across all neurons (\cref{fig:characterization}A).
For multiple configurations on this spectrum, we then continued training with hardware in the loop for one epoch.

Within that single epoch, the network adapted to the changed parametrization of the substrate (\cref{fig:mnist_decalibration}):
Depending on the strength of decalibration, a strong loss of performance could be observed before re-training.
After one epoch, a test accuracy of \SI{96.07}{\percent} was restored (decalibration factor: \num{1.0}), representing only a slight drop down from \SI{96.31}{\percent} for the calibrated network.

In the uncalibrated state, synaptic input amplitudes differed by up to a factor of four.
This can be interpreted as a reduction of the effective weight resolution from 6 bit to only 4 bit.

We replicated this reduction of resolution to 4 bit in software and observed a reduced accuracy of \SI{96.99}{\percent} compared to the original performance of \SI{97.36}{\percent}.
Our results therefore coincide with the expectations.

%% file: tex/4_discussion.tex
\section{Discussion}\label{sec:discussion}

In this publication, we have shown that \BrainScaleS{2} can be successfully used for vector-matrix multiplication, especially in the context of deep convolutional neural networks.
We have presented a set of calibration mechanisms to set up the analog system and equalize fixed-pattern variations of the computational units.
Improvements upon a pre-trained and directly transferred network could be reached through training with hardware in the loop, compensating for remaining imperfections.
Since calibration data can be generated once and then be re-used for multiple networks, calibration can still prove valuable compared to task-specific training.

Compared to the same network evaluated in software, we reached state-of-the-art classification performance on the MNIST dataset.
Shortcomings of the current hardware generation result in a reduced multiplication throughput and therefore far from optimal energy efficiency.
The upcoming revision of the system addresses all the underlying issues and promises efficiencies and runtimes improved by a factor of \numrange{100}{1000}.

%% file: tex/5_contributions.tex
\section{Contributions}

J.~Weis developed calibration routines, conducted the presented experiments and evaluations and wrote the initial manuscript.
P.~Spilger is the main developer of the software extensions providing support for BrainScaleS' non-spiking operation mode.
S.~Billaudelle designed neuron and synapse driver circuits and contributed to commissioning of the chip.
Y.~Stradmann contributed to hardware design and commissioning and gave conceptual advice.
A.~Emmel contributed to experiment code.
E.~Müller is the lead developer and architect of the \BrainScaleS{2} software stack.
C.~Mauch and  O.~Breitwieser contributed to the software architecture and implementation.
A.~Grübl was responsible for chip assembly and implemented the digital front- and backend.
J.~Ilmberger contributed to host-side communication infrastructure.
V.~Karasenko is the main developer of the FPGA firmware and developed the communication infrastructure between FPGA and ASIC.
M.~Kleider contributed to FPGA firmware development as well as initial commissioning of the system.
K.~Schreiber designed and implemented the CADC and the physical ASIC test setup.
J.~Schemmel is the lead designer and architect of the \BrainScaleS{2} neuromorphic system.
All authors discussed and contributed to the manuscript.